\documentclass{article} % For LaTeX2e
\usepackage{arxiv2025,times}

\usepackage{amsmath,amsfonts,bm}

\DeclareMathOperator{\defeq}{\stackrel{\text{def}}{\;=\;}}

\def\eqref#1{equation~\ref{#1}}
\def\1{\bm{1}}

\DeclareMathAlphabet{\mathsfit}{\encodingdefault}{\sfdefault}{m}{sl}
\SetMathAlphabet{\mathsfit}{bold}{\encodingdefault}{\sfdefault}{bx}{n}

\usepackage{hyperref}
\usepackage{url}
\usepackage{graphicx}
\usepackage{algorithm}
\usepackage{algpseudocode}
\usepackage{xcolor}
\usepackage{mdframed}
\usepackage{booktabs}

\title{ZIP-FIT: Embedding-Free Data Selection via Compression-Based Alignment}

\author{%
  \textbf{Elyas Obbad}\textsuperscript{1}, \textbf{Iddah Mlauzi}\textsuperscript{1}, \textbf{Brando Miranda}\textsuperscript{1}, \\
  \textbf{Rylan Schaeffer}\textsuperscript{1}, \textbf{Kamal Obbad}\textsuperscript{3}, \textbf{Suhana Bedi}\textsuperscript{2}, \textbf{Sanmi Koyejo}\textsuperscript{1} \\
  \texttt{\{eobbad, iddah, brando9, sanmi\}@cs.stanford.edu} \\
  \textsuperscript{1}Department of Computer Science, Stanford University\\
  \textsuperscript{2}Department of Biomedical Data Science, Stanford School of Medicine\\
  \textsuperscript{3}Stanford Biophysics Program, Stanford School of Medicine\\
}

\iclrfinalcopy % Uncomment for camera-ready version, but NOT for submission.
\begin{document}

\maketitle

\begin{abstract}
    Data selection is crucial for optimizing language model (LM) performance on specific tasks, yet most existing methods fail to effectively consider the target task distribution. 
    Current approaches either ignore task-specific requirements entirely or rely on approximations that fail to capture the nuanced patterns needed for tasks like Autoformalization or code generation.
    Methods that do consider the target distribution often rely on simplistic, sometimes noisy, representations, like hashed n-gram features, which can lead to collisions and introduce noise.
    We introduce \texttt{ZIP-FIT}, a data selection framework that uses \texttt{gzip} compression to directly measure alignment between potential training data and the target task distribution.
    Our key insight is that compression-based similarity captures both syntactic and structural patterns relevant to the target task, enabling more precise selection of truly task-relevant data.
    In extensive evaluations on Autoformalization and Python code generation, \texttt{ZIP-FIT} significantly outperforms leading baselines like DSIR and D4. Models trained on \texttt{ZIP-FIT}-selected data achieve their lowest cross-entropy loss up to 85.1\% faster than baselines, demonstrating that better task alignment leads to more efficient learning. In addition, \texttt{ZIP-FIT} performs selection up to 65.8\% faster than DSIR and two orders of magnitude faster than D4. Notably, \texttt{ZIP-FIT} shows that smaller, well-aligned datasets often outperform larger but less targeted ones, demonstrating that a small amount of higher quality data is superior to a large amount of lower quality data. Our results imply that task-aware data selection is crucial for efficient domain adaptation, and that compression offers a principled way to measure task alignment. By showing that targeted data selection can dramatically improve task-specific performance, our work provides new insights into the relationship between data quality, task alignment, and model learning efficiency.
\end{abstract}

\section{Introduction}
\label{sec:introduction}

Choosing training data is crucial for the performance of language models (LMs) in both general-purpose and domain-specific applications \citep{brown2020languagemodelsfewshotlearners, dontstoppretrainingadapt, trainingcomputeoptimallargelanguage}. 
To date, most research on data curation has focused on creating diverse pre-training datasets to enhance model performance across a wide range of tasks \citep{sorscher2022beyond,dsir, d4, semdedup, doremi,lee2023beyond,wettig2024quratingselectinghighqualitydata,penedo2024finewebdatasetsdecantingweb,li2024datacomp,sachdeva2024train}, and while these methods have been demonstrated to work well for general pre-training, they fall short in domain-specific fine-tuning, where data relevance is crucial. This raise a key question: 
\textit{How should we, in a general purpose manner, effectively select fine-tuning data for a domain-specific target task?}

One approach is to train binary classifiers to identify relevant data. 
For example, a mathematical language model called DeepSeekMath \citep{deepseekmath} utilized OpenWebMath \citep{openwebmathopendatasethighquality}, a compilation of high-quality mathematical texts, to train a FastText classifier % \citep{enrichingWordVectorsWithSubwordInfo}
to retrieve analogous texts from the Web \citep{enrichingWordVectorsWithSubwordInfo}. 
Although effective, this method relies on the availability of large and well-annotated data sets, something that is often missing in niche tasks where relevant data are scarce. 
\begin{figure*}[t]
  \centering
  {\includegraphics[width=\textwidth]{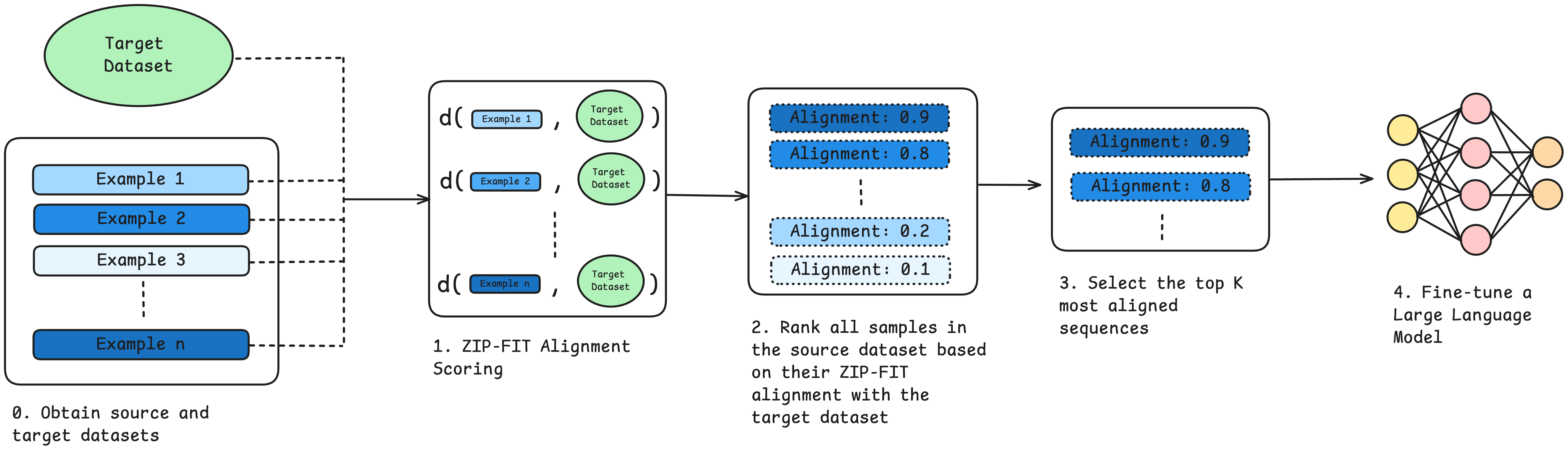}}
  \caption{\textbf{\texttt{ZIP-FIT} selects task-specific data for efficient finetuning.} 
  (0) Obtain both the source and target datasets. 
  (1) Calculate \texttt{ZIP-FIT} Alignment of each source example with the target dataset using \texttt{gzip} compression. 
  (2) Rank all source examples based on these alignment scores. 
  (3) Select the top-K most aligned examples for fine-tuning. 
  (4) Fine-tune a large language model using the selected top-K examples to improve performance on the target task.}
  \label{fig:zip-fit-vs-baselines}
\end{figure*}
\begin{figure*}[t]
  \centering
  {\includegraphics[width=1.0\textwidth]{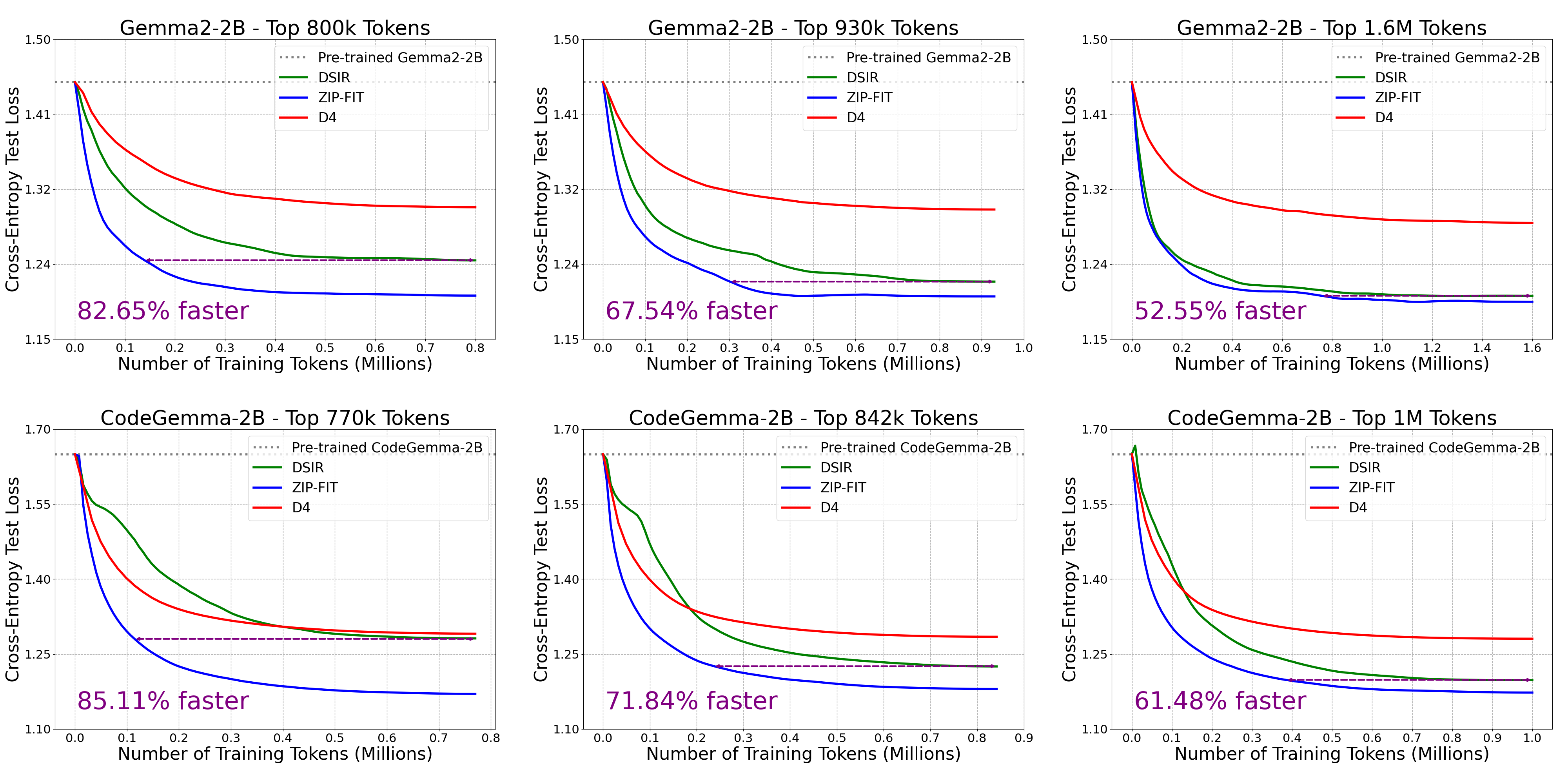}}
  \caption{\textbf{Code Generation: \texttt{ZIP-FIT} accelerates cross-entropy loss reduction, even in code-specialized models like CodeGemma-2B.} The plots show cross-entropy test loss versus the number of training tokens for Gemma2-2B (top row) and CodeGemma-2B (bottom row) across different token selection sizes. \texttt{ZIP-FIT} (blue) consistently reduces loss faster than DSIR (green) and D4 (red), achieving up to $85.11\%$ speed improvement at lower token counts. These results demonstrate \texttt{ZIP-FIT}'s efficiency in data selection for fine-tuning models on code-geneation tasks.
  }
  \label{fig:code-all-models}
\end{figure*}
Another common approach is to use neural embeddings to measure the similarity between data points and a reference corpus \citep{efficientcontinualpretrainingbuilding}. 
Although this improves relevance, embedding-based methods are computationally expensive and sensitive to the choice of embedding space \citep{sgptgptsentenceembeddings}. 
Alternatively, DSIR (Data Selection via Importance Resampling) \citep{dsir} utilizes unigrams and bigrams to select data points without the need for pre-trained embeddings, with the aim of matching the hashed n-gram distributions of the target data. Although DSIR is effective in capturing direct word correlations, it may not capture structured patterns of syntax that unfold across sentences or paragraphs, such as nested function calls in code or embedded clauses in formal language translation \citep{deMoura2015lean}. Additionally, the hashing introduces noise due to collisions. These shortcomings highlight the need for alternative data selection strategies better suited to domain-specific tasks.

To address these challenges, we propose \texttt{ZIP-FIT}, a novel data selection framework that leverages the classic compression algorithm \texttt{gzip}.
Recent research suggests that language modeling and data compression are fundamentally equivalent tasks \citep{languagemodelingcompression}, and the intelligence of large language models (LLMs) is closely related to their ability to compress external corpora \citep{compressionrepresentsintelligencelinearly}. 
This insight suggests that compression algorithms can encode information in ways similar to neural networks. 
For example, \citet{jiang-etal-2023-low} found that the use of normalized compression distances for text classification outperformed traditional neural embeddings. 
Inspired by this, \texttt{ZIP-FIT} selects aligned training data with a target data set based on compression-based alignment, providing a lightweight and embedding-free method for selecting high-quality data.

% We evaluated \texttt{ZIP-FIT} across two distinct domains: Autoformalization and Python code generation. 
% We show that \texttt{ZIP-FIT} outperforms existing data selection frameworks and consistently improves model performance, particularly when evaluating cross-entropy test loss for target domains. 
% Our experiments reveal that smaller, well-aligned datasets, chosen through \texttt{ZIP-FIT}, lead to faster convergence and superior performance compared to the use of larger, less aligned data sets, underscoring the importance of quality over quantity. 
We evaluated \texttt{ZIP-FIT} in two domains: Autoformalization and Python code generation. 
\texttt{ZIP-FIT} outperforms existing data selection methods, consistently improving model performance cross-entropy test loss. 
Smaller, well-aligned datasets selected by \texttt{ZIP-FIT} lead to faster convergence and better performance than larger, less aligned datasets, highlighting the importance of data quality.

Our \textbf{contributions} are as follows:
\begin{enumerate}
    \item \textbf{Methodology:} The introduction of \texttt{ZIP-FIT}, an embedding-free data selection method based on \texttt{gzip} compression.
    \item \textbf{Superior Performance:} \texttt{ZIP-FIT} consistently outperforms leading baselines (DSIR, D4) in Autoformalization and Python code generation, achieving up to 85.1\% faster convergence and lower cross-entropy loss.
    \item \textbf{Computational Efficiency:} \texttt{ZIP-FIT} is computationally efficient, running up to 65.8\% faster than DSIR. This makes it scalable for low-resource environments without compromising performance.
\end{enumerate}

\section{\texttt{ZIP-FIT}: an Embedding-Free Data Selection Algorithm via Compression-Based Alignment for LM Fine-Tuning}
\begin{figure*}[t]
    \centering
    {\includegraphics[width=\textwidth]{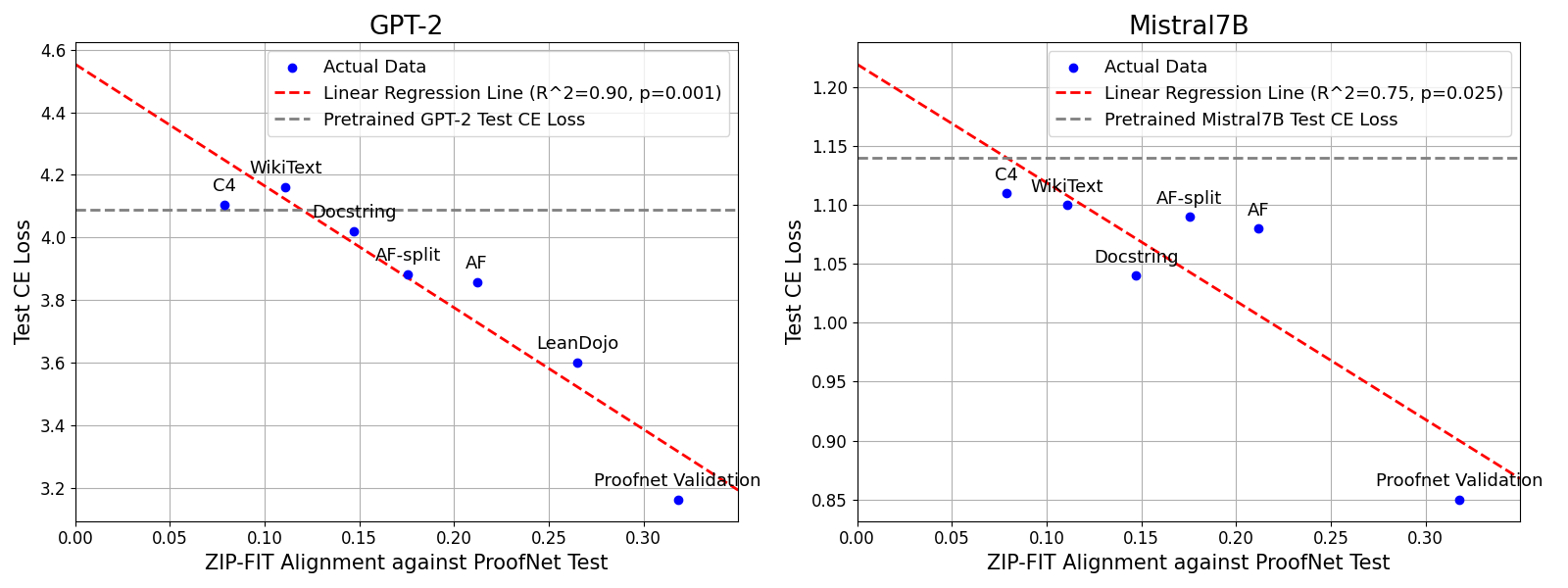}}
    \caption{\textbf{Higher \texttt{ZIP-FIT} alignment correlates with lower cross-entropy loss.} The relationship between \texttt{ZIP-FIT} alignment and cross-entropy (CE) loss for (a) GPT-2 trained on 22k tokens ($R^2 = 0.90, p=0.001$) and (b) Mistral7B trained on 22k tokens ($R^2 = 0.75, p=0.025$). Each point represents a dataset, with its position reflecting the dataset’s \texttt{ZIP-FIT} alignment score against the ProofNet test set and the resulting CE loss. The dashed red line indicates the linear regression fit, while the dashed grey line shows the pretrained CE loss. Higher alignment scores correspond to lower CE losses, demonstrating that training on better aligned data yields better performance.}
    \label{fig:ce_vs_gzip_align}
\end{figure*}
Before introducing \texttt{ZIP-FIT}, it is essential to understand the desired attributes of effective data selection algorithms. Ideally, such algorithms should be performant, computationally economical, fast, scalable, and designed to improve the efficiency of model training. These characteristics ensure that the data filtering process can be applied broadly and effectively in various machine learning contexts, particularly when computational resources are limited. By setting these criteria, we can better appreciate the innovations \texttt{ZIP-FIT} introduces in the realm of data selection.

\subsection{Background}

\textbf{\texttt{gzip} compression}: uses two main techniques for compression, LZ77 and Huffman coding. 
Together, these methods compress sequences by exploiting repeated patterns in the data. 
LZ77 works by identifying repeated substrings and replacing them with references to their earlier occurrences.
Huffman coding further compresses the data by assigning shorter binary codes to more frequent symbols, optimizing the overall length of the compressed text.
For more details, see Appendix~\ref{app:sec:gzip_compression_details}.

\textbf{AutoFormalization (AF):} refers to the task of translating natural language mathematical statements into a formal mathematical programming language, like Lean4 \cite{deMoura2015lean}. 
This process requires precise understanding and representation of mathematical formal syntax, making the selection of well-aligned training data crucial for effective model training.

\subsection{\texttt{ZIP-FIT} Algorithm}

\textbf{Setup:} Given a set of examples \(\{x'_1, x'_2, \ldots, x'_n\}\) from a target distribution \(p\) 
% (e.g., $n = 185$ for ProofNet's validation split) 
and a large source dataset \(\{x_1, x_2, \ldots, x_N\}\) from an arbitrary distribution \(q\), \texttt{ZIP-FIT} aims to select a subset of K examples (where \(K \ll N\)) from \(q\). The selected subset is used for model training, in order to improve performance for tasks associated with \(p\). This approach is intended to maximize the efficacy and efficiency of model training by focusing on the most relevant data samples.
%\rylan{what is a domain? a distribution? right now, this is unclear} ADDRESSED \rylan{I feel like this goal is written inaccurately. I think our goal is to improve performance at $p$, and to do this, we want to determine how we can use this large dataset to improve performance at $p$ }

\textbf{Method:} \texttt{ZIP-FIT} uses \texttt{gzip} compression as a metric to measure the alignment of each example in \(q\) with the target \(p\), focusing on capturing patterns and redundancies.

To address the challenge of selecting highly aligned data, we propose the \texttt{ZIP-FIT} algorithm:

\begin{algorithm}
\caption{\texttt{ZIP-FIT} Data Selection Algorithm}
\begin{algorithmic}[1]
\State \textbf{Input:} A large source dataset \(D = \{x_1, x_2, \dots, x_N\}\) from distribution \(q\), target examples \(\{x'_1, x'_2, \dots, x'_n\}\) from distribution \(p\).
\State \textbf{Output:} A subset of K examples from \(D\) that improve performance at \(p\).
\For{\(i = 1\) to \(N\)}
    \State Compute alignment for each sample \(x_i \in D\) with each target example 
    \Statex \hspace{\algorithmicindent}\(x'_j \in \{x'_1, x'_2, \dots, x'_n\}\) using Normalized Compression Distance:
    \Statex \[ \text{NCD}(x_i, x'_j) \defeq \frac{C(x_i \oplus x'_j) - \min(C(x_i), C(x'_j))}{\max(C(x_i), C(x'_j))} \]
    \Statex \hspace{\algorithmicindent}where \(C(x)\) represents the compressed size of sequence \(x\) and \(\oplus\) denotes concatenation.
    \State Compute the average \texttt{ZIP-FIT} alignment for each \(x_i\):
    \[ \text{\texttt{ZIP-FIT}-Alignment}(x_i) \defeq 1 - \frac{1}{n} \sum_{j=1}^{n} \text{NCD}(x_i, x'_j) \]
\EndFor
\State Select the top-K examples from \(D\) based on the highest alignment scores.
\end{algorithmic}
\end{algorithm}

% Decision: Moving to appendix for now
% \subsection{Why Use Compression?}

% Compression algorithms, such as \texttt{gzip}, provide a computationally efficient way to detect patterns and minimize redundancy in data. 
% \paragraph{Limitations of n-grams:} Many traditional methods, including hashed n-grams, focus on capturing immediate textual correlations by simplifying text into discrete, fixed-size buckets. 
% Although these techniques are computationally efficient, they may not adequately capture syntactic or structural relationships within the data. Additionally, the introduce noise due to collisions during hashing.

% \paragraph{Challenges with Neural Embeddings:} Neural embeddings offer a powerful tool for capturing semantic relationships, but they come with significant computational costs. These embeddings are typically pre-trained on large corpora and fine-tuned for specific tasks, which requires substantial resources. Given the scalability challenges of embedding-based methods, we conjecture that a simpler method like compression can provide a more scalable and resource-efficient alternative.

% We hypothesize that compression -- in this case \texttt{gzip}, but perhaps a different compression algorithm --serves as a strong proxy for capturing syntactic and structural relationships in textual sequences. 
% \texttt{gzip}’s ability to compress data based on redundancy minimization can be leveraged as a metric to align text with a target distribution. 
\section{Higher alignment interventionally achieves better model performance }

\textbf{Experiment:} We validate compression as an alignment metric by evaluating the impact of a model fine-tuned on more \texttt{ZIP-FIT}-aligned data with a target task and the corresponding cross-entropy (CE) loss.  
We chose ProofNet (test) as the target benchmark and then fine-tuned GPT-2 \cite{radford2019language} and Mistral7B \cite{mistral7b} ) LMs on datasets with varying \texttt{ZIP-FIT} alignment. 

\textbf{Results:} Figure~\ref{fig:ce_vs_gzip_align} shows a strong negative correlation ($R^2$ of 0.90) between \texttt{gzip} alignment scores and CE loss for GPT-2 and 0.75 for Mistral7B. 
This implies that data alignment plays a crucial role in improving model performance 
%\rylan{Are we controlling for number of toekns here? If so, please say that. If not, how is this a valid comparison?}. 
We control for the number of tokens in each dataset, setting it to 100k tokens, except for datasets that do not contain this many tokens (e.g., ProofNet validation set). 
Data sets such as LeanDojo \cite{yang2023leandojo} and the ProofNet validation set, which exhibit high alignment scores, resulted in significantly lower CE loss compared to less aligned data sets such as C4 and WikiText (\cite{raffel2020exploring}, \cite{merity2016pointer}).
Data sets with high alignment like LeanDojo \cite{yang2023leandojo} and the ProofNet (validation) resulted in a significantly lower CE loss compared to less aligned data sets like C4 and WikiText (\cite{raffel2020exploring}, \cite{merity2016pointer}).

\section{Higher Alignment leads to more efficient training}

\textbf{Experiment:} We fine-tuned GPT-2 (124M) and Mistral7B for the AutoFormalization task using different datasets scored with \texttt{ZIP-FIT} alignment. We used ProofNet (test) for the evaluation.
The curves represent different datasets with varying alignment to the target domain (ProofNet validation). 
% Higher alignment values indicate a more targeted data selection.

\textbf{Results:} More aligned data reduces CE loss quickest, as shown by the steep decline for high-alignment datasets. 
This is most evident as ProofNet (validation). % which has an average alignment score of 0.32. 
Less aligned data require significantly more tokens to achieve similar performance. 
This demonstrates that targeted data selection with \texttt{ZIP-FIT} accelerates fine-tuning and improves performance, reducing computational costs. 

\begin{figure*}[t]
  \centering
  {\includegraphics[width=1.0\textwidth]{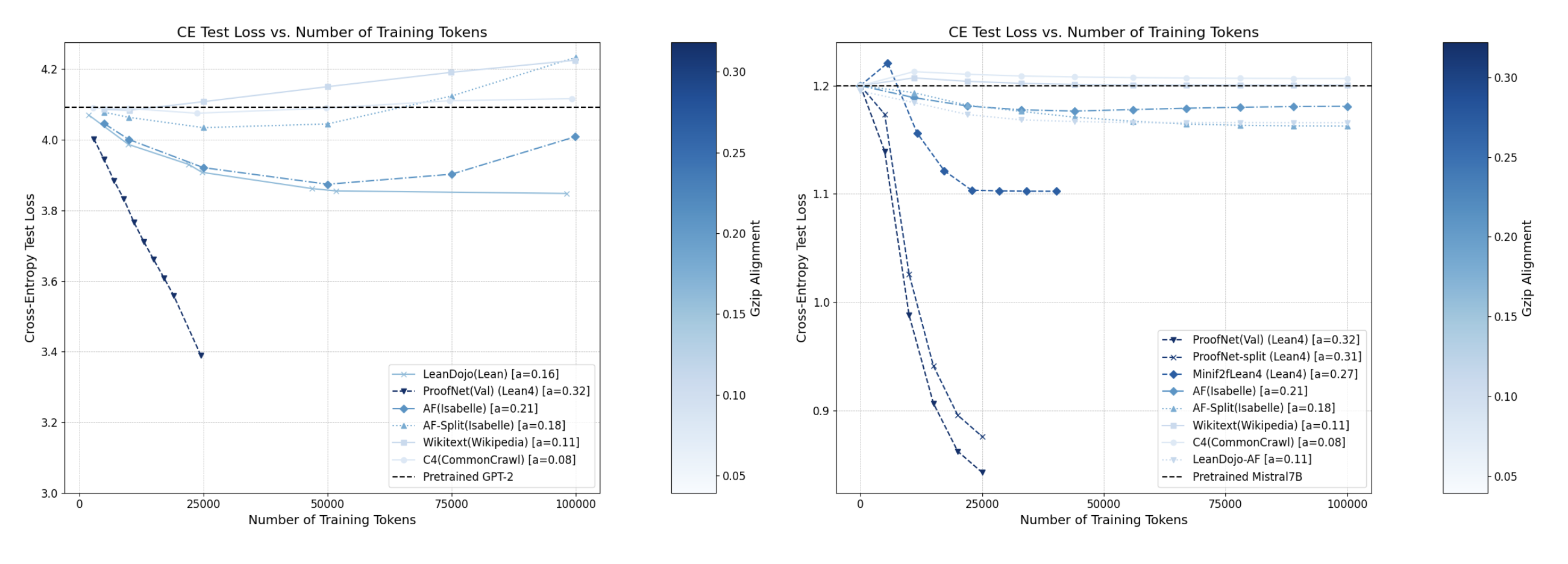}}
  \caption{\textbf{Highly aligned data lowers cross-entropy loss more efficiently.} 
  The x-axis shows the number of training tokens, and the y-axis represents the cross-entropy (CE) test loss on the ProofNet test set. Different curves correspond to datasets filtered by different alignment scores, indicating their relevance to the target domain. The most aligned data reduce Test CE loss significantly faster than less aligned data. The left panel depicts results using GPT-2, and the right panel uses Mistral7B, demonstrating that using highly aligned data not only accelerates training but also achieves better model performance, validating the effectiveness of \texttt{ZIP-FIT} for data selection in fine-tuning.}
  \label{fig:more-alignment-lower-loss}
\end{figure*}

\section{Comparative Evaluation of \texttt{ZIP-FIT} for Efficient Fine-Tuning}
\begin{figure*}[]
  \centering
  {\includegraphics[width=\textwidth]{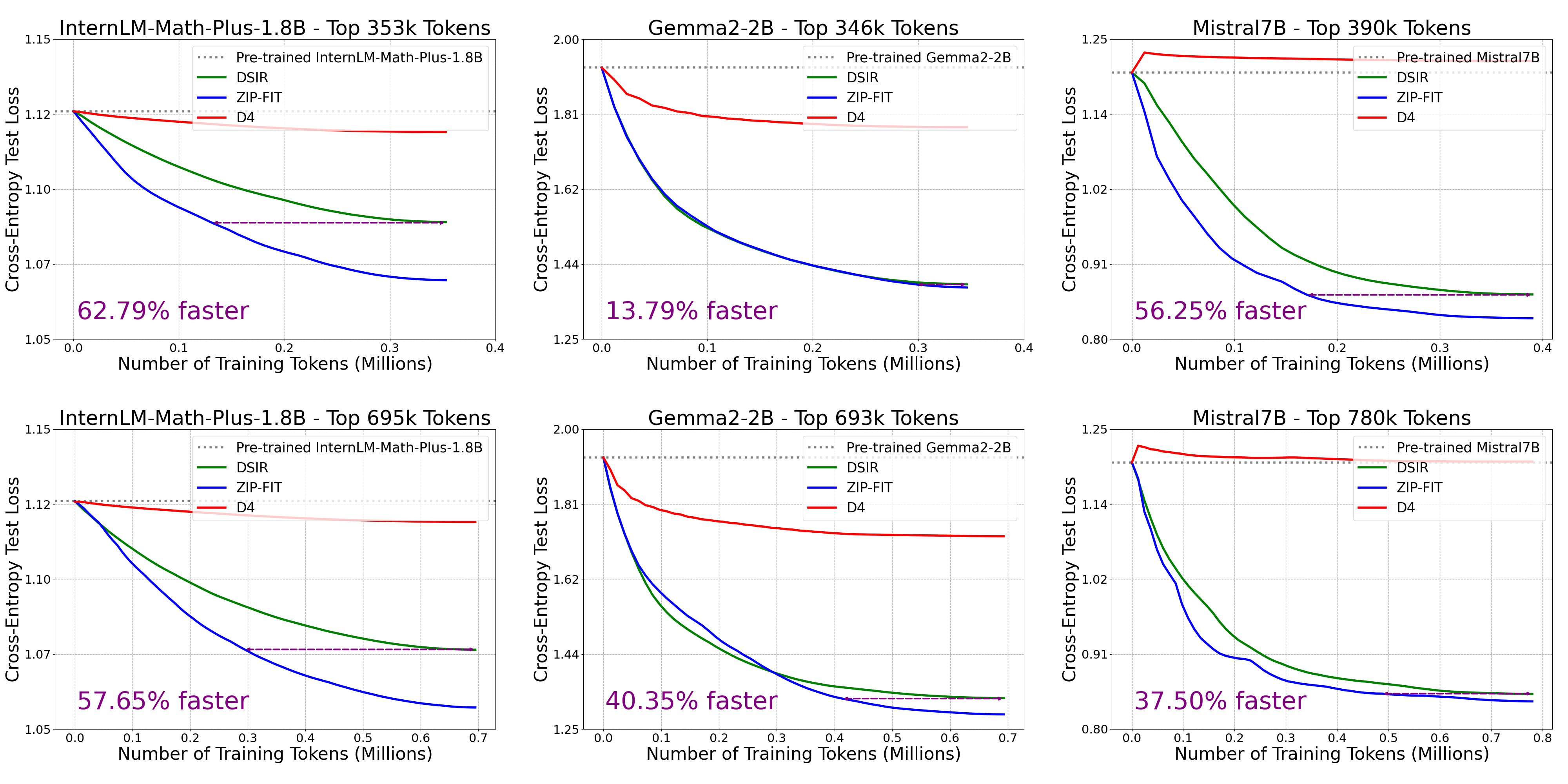}}
  \caption{\textbf{AutoFormalization: \texttt{ZIP-FIT} consistently achieves lower test loss more quickly than \texttt{D4} and \texttt{DSIR}, demonstrating its efficiency in data selection.} The plots show cross-entropy test loss versus the number of training tokens for three models (InterLM-Math-Plus-1.8B, Gemma2-2B, and Mistral7B) across different token selection sizes. \texttt{ZIP-FIT} (blue line) consistently outperforms both \texttt{DSIR} (green line) and D4 (red line) across all model and token size configurations, highlighting its ability to process data more efficiently. The percentage labels in each plot indicate the relative speedup of \texttt{ZIP-FIT} over \texttt{DSIR} in reaching the lowest cross-entropy loss, reinforcing the method's scalability and adaptability for domain-specific fine-tuning. %\rylan{(1) this figure is small and hard to read. It also appears to not stretch across the page? (2) \rylan{For certain subfigures, where green and purple never match blue, you might also want to add a vertical line saying x\% better (to completement the \% faster).}}
  }
  \label{fig:af-all-models}
\end{figure*}
We evaluate \texttt{ZIP-FIT} on two domain-specific tasks: \textit{Autoformalization} and \textit{Python Code Generation}. 
% Our goal is to demonstrate \texttt{ZIP-FIT}'s ability to select data that leads to better fine-tuning performance compared to leading baselines.
Our goal is to show \texttt{ZIP-FIT}'s data selection leads to superior fine-tuning.

\subsection{AutoFormalization}
\label{app:subsec:autoformalization_dataset}

\paragraph{Experiment:} Our source dataset comprised approximately 185,000 sequences from LeanDojo, Proof-Pile 2, C4, and WikiText \cite{yang2023leandojo, azerbayev2024llemmaopenlanguagemodel}. 
For details, refer to Appendix \ref{app:subsec:dataset_composition_af}. 
Alignment was computed using \texttt{ZIP-FIT}, DSIR and D4 with the ProofNet's validation split (our target distribution).
For a fair comparison, we did not modify how any of the methods rank sequences. 
To compare each method, we selected the n sequences ranked highest for several values of n (353k, 695k tokens, etc.). 
For each selected data set at each value of n we fine-tune InterLM-Math-Plus-1.8B, Gemma2-2B, and Mistral7B (\cite{ying2024internlmmathopenmathlarge}; \cite{team2024gemma}).
Performance was evaluated using the CE loss ProofNet's test split.

\textbf{Results:} 
Figure~\ref{fig:af-all-models} shows that \texttt{ZIP-FIT} significantly outperforms DSIR and D4 in reducing cross-entropy (CE) loss across all token selection sizes (353k, 695k).
The steep decline in the blue curves (\texttt{ZIP-FIT}) highlights its ability to achieve faster convergence, resulting in up to 62.79\% improvements in convergence speeds compared to DSIR. 
Notably, \texttt{ZIP-FIT} demonstrates up to a 65.8\% faster data selection process than DSIR. 
% \texttt{ZIP-FIT} demonstrates efficiency in leveraging highly-aligned data, as seen in its strong performance on InterLM-Math-Plus-1.8B, a model already optimized for mathematical tasks. 
An interesting observation is \texttt{ZIP-FIT}'s efficiency in selecting highly-aligned data and improving even specialized mathematical models like InternLM-MAath-Plus-1.8B. 
While one might expect diminished returns on a model already adept in a related domain, the improvements suggest that the model still benefits from the AutoFormalization data. 
% This rapid reduction underlines \texttt{ZIP-FIT}'s efficiency in utilizing highly-aligned data, especially notable in its superior performance on InterLM-Math-Plus-1.8B, which is already optimized for mathematical data. 
% This is surprising because one would expect little to no gains on models that already know how to do the target task. 
% This advantage emphasizes that \texttt{ZIP-FIT} not only accelerates learning, but also enhances the effectiveness of fine-tuning, even on models predisposed to mathematics, reinforcing its utility for AutoFormalization. 
Similar results were observed at other token counts, as detailed in Appendix \ref{sec:more-af-plots}. 

\subsection{Code Generation}

\paragraph{Experiment:} 
We conducted code generation experiments using \texttt{ZIP-FIT}, DSIR, and D4 to select data from a mix of sources:
MBPP (\cite{austin2021programsynthesislargelanguage}, Python docstrings, Proof-Pile 2, C4, WikiText. 
The latter two are included to study whether the data selection methods considered are robust to misaligned data. 
For details, refer to Appendix \ref{app:subsec:dataset_composition_code}.
The datasets were utilized to fine-tune both CodeGemma-2B and Gemma2-2B models, with the focus on translating function signatures and docstrings into executable Python code. 
For the selection process, we used HumanEval for validation and a separate hold-out portion for final testing. 
% We varied the top k selections to explore different dataset sizes.
For a fair comparison, we did not modify how any of the methods rank sequences. 
To compare each method, we selected the n sequences ranked highest for several values of n (800k, 930k, 1.6M tokens, etc.). 

\paragraph{Results:} Across all tested n values, \texttt{ZIP-FIT} consistently outperformed DSIR and D4 in reducing cross-entropy loss, demonstrating faster and more effective fine-tuning. 
In particular, the CodeGemma-2B model, already optimized for code-related tasks, showed the most improvements with \texttt{ZIP-FIT}, confirming its ability to select highly relevant and beneficial training data. 
Rapid loss reduction under \texttt{ZIP-FIT} emphasizes its efficiency, especially noted in its 25\% faster data processing compared to DSIR. 
Most notably, the flattening of the DSIR and D4 curves indicate diminishing returns, suggesting that additional tokens would not achieve the performance of \texttt{ZIP-FIT}. 
In general, these findings emphasize that \texttt{ZIP-FIT} accelerates model training and optimizes resource usage, making it a superior choice for code generation tasks.

\section{Impact of Data Misalignment on Model Performance}

Existing research showed that data alignment plays a critical role in improving model performance and learning efficiency for downstream tasks. In this section, we explore how misalignment in data can affect model performance and how \texttt{ZIP-FIT} addresses this issue with data selection.

\begin{figure*}[t]
  \centering
  % {\includegraphics[width=0.7\textwidth]{./plots/misalignment.png}}
  {\includegraphics[width=1\textwidth]{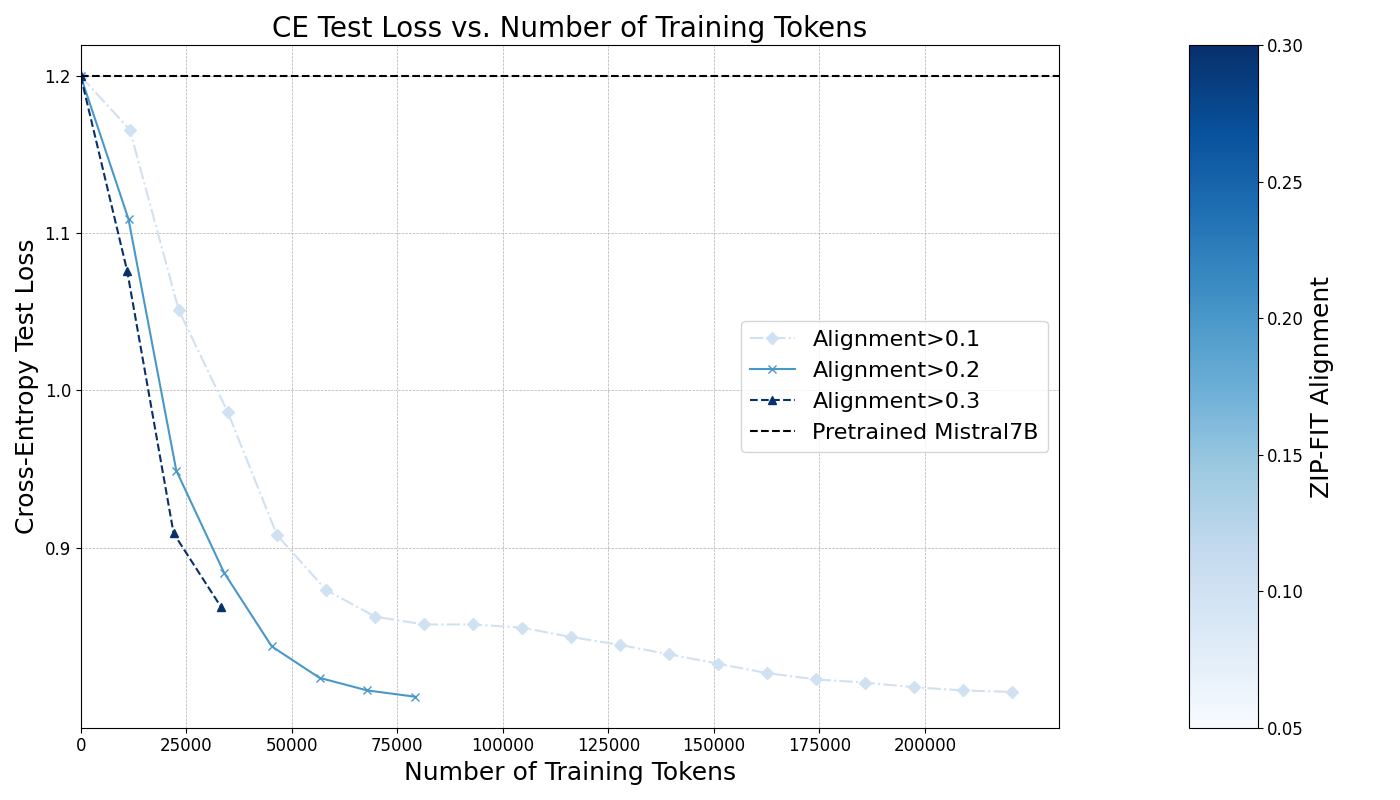}}
  \caption{\textbf{Selective data filtering with \texttt{ZIP-FIT} allows us to achieve better cross-entropy test loss faster than training on all the data, resulting in improved performance and efficiency.} 
  The x-axis represents the number of training tokens, while the y-axis shows the cross-entropy test loss. 
  The curves represent models fine-tuned (FT) on datasets filtered by varying alignment thresholds ($>$ 0.1, $>$ 0.2, $>$ 0.3).
  The dashed line indicates the baseline performance of the pretrained Mistral7B model. 
  Training on data filtered with higher alignment thresholds leads to superior performance, demonstrating the effectiveness of removing misaligned data in fine-tuning.
  }
  \label{fig:misaligned-data}
\end{figure*}

\textbf{Experiment:} We fine-tuned the Mistral7B model on the same source dataset we used for the AutoFormalization experiment (see Appendix \ref{app:subsec:autoformalization_dataset}), filtering data with \texttt{ZIP-FIT} at different alignment thresholds ($>$0.1, $>$0.2, $>$0.3). 
Each threshold creates a progressively more aligned dataset, where the $>$0.3 dataset is the most aligned, and the $>$0.2 dataset is a superset of the $>$0.3 dataset, including less aligned data. 
Similarly, the $>$0.1 dataset is a superset of both $>$0.2 and $>$0.3.
Figure~\ref{fig:misaligned-data} shows CE test loss (y-axis) versus the number of training tokens (x-axis). 

\textbf{Results:} \texttt{ZIP-FIT} selected data achieves lower CE loss faster than training on all data (Figure~\ref{fig:misaligned-data}), showing improved performance and efficiency. 
Higher alignment thresholds result in a steeper loss reduction, confirming that \textit{filtering out misaligned data enhances fine-tuning}. 
Misalignment can introduce noise and irrelevant patterns, which we hypothesize require more training data and computational resources to overcome. 
Applying higher alignment thresholds, \texttt{ZIP-FIT} ensures that only the most relevant examples are used for training. 
This targeted selection leads to a more efficient learning process as evidenced by the sharper decline in CE loss for higher alignment thresholds. 
Such efficiency is crucial in scenarios where computational resources are limited or costly.

% \textbf{Theoretical Implications:} The observed trends underscore the theoretical implications of information theory in machine learning, where reducing the entropy or randomness in the input data directly contributes to better model performance. This aligns with the concept that a cleaner, more relevant dataset effectively reduces the hypothesis space that the model needs to explore during training.

\textbf{Practical Considerations:} For practitioners, these results suggest that investing in better data curation and alignment tools can significantly cut down the cost and time of model training without compromising performance. 
It also highlights the potential pitfalls of using large, uncurated datasets that might slow down the learning process or lead to poorer generalization on specific tasks.

\textbf{Future Directions:} Could explore adaptive alignment thresholds based on real-time validation CE, potentially automating the selection process to optimize both speed and accuracy during training. 
 
Filtering out misaligned data accelerates fine-tuning and reduces computational overhead, confirming its performance gains and computational efficiency as outlined in our contributions.
% utility in low-resource settings.

\section{Related Works}

\textbf{Curating pre-training data} often involves using classifiers to filter high-quality data from large corpora like Common Crawl, as done for models like GPT-3 and PaLM2 \citep{brown2020languagemodelsfewshotlearners, palm2technicalreport2023, deepseekmath}. While effective, this process requires significant computational resources and large volumes of curated data. In contrast, \texttt{ZIP-FIT} efficiently selects relevant data without relying on external models, making it especially useful in data-scarce environments.

\textbf{Deduplication} techniques, such as SemDeDup \citep{semdedup} and D4 \citep{d4}, improve data efficiency by removing duplicate or semantically similar examples using embedding-based clustering. However, these methods are computationally expensive and not tuned to the target task. \texttt{ZIP-FIT} is embedding-free and task-aware, making it both scalable and more effective at selecting relevant data.

\textbf{Mixture weights} are essential when drawing from multiple domains, as they influence the performance of language models \citep{glamefficientscalinglanguage, dsir}. DoReMi (Domain Reweighting with Minimax Optimization) \citep{doremi} proposes a reweighting strategy suitable for handling diverse target distributions, but it primarily focuses on adjusting weights at the domain level. Adapting it to select individual data points for specific target distributions would require substantial modifications to its foundational algorithm. 
One potential approach would be to effectively transform each data point into a 'mini-domain,' a process that would stray significantly from DoReMi's original purpose and scope. 
Therefore, we did not use DoReMi in our comparisons because it does not directly address the fine-grained selection needs that \texttt{ZIP-FIT} fulfills. 

\textbf{Autoformalization} refers to the process of translating natural language mathematics into formal language \citep{Wang_2020, autoformalizationlargelanguagemodels}, which is advantageous because formal proofs can be verified for correctness. 
However, the ability of current models to autoformalize text is limited by the scarcity of human-curated formal data. \texttt{ZIP-FIT} provides a framework for selecting the most relevant data, ensuring that models are trained on aligned datasets that enhance their performance. 

\textbf{Compression: } 
\citep{pandey2024gzippredictsdatadependentscaling} demonstrates that gzip-compressibility predicts how language model scaling laws shift with data complexity, challenging the data-agnostic assumptions of Chinchilla-style scaling laws.
\citep{jiang2022gzip} use \texttt{gzip}-based compression distance with kNN for zero-shot text classification, outperforming BERT on some datasets.
\citep{languagemodelingcompression} show that language modeling is equivalent to compression, where even \texttt{gzip} can define predictive distributions via coding length.
\citep{yoran2025kolmogorovtestcompressioncode} propose the KoLMogorov Test, showing that \texttt{gzip} remains a strong baseline for code-based compression, outperforming pretrained LLMs on real data.
 show that \texttt{gzip} compression ratio is a fast, effective proxy for text diversity, capturing key repetition patterns in LLM outputs and aligning with slower lexical metrics e.g., self-BLEU.

\section{Limitations}
While \texttt{ZIP-FIT} provides a computationally efficient method for data selection, it has several limitations. 
First, the \texttt{gzip} compression-based alignment may not fully capture nuanced semantic relationships that dense representations can, potentially affecting its effectiveness for complex domains like natural language understanding, where paraphrasing is important. 
Second, \texttt{ZIP-FIT}’s reliance on \texttt{gzip} means that its performance could vary depending on the nature of the textual data, particularly in highly diverse datasets where compression gains are less apparent.

\section{Discussion and Future Work}

\texttt{ZIP-FIT} introduces an efficient, embedding-free approach for data selection in language model fine-tuning. 
By leveraging compression algorithms to capture redundancies in data, \texttt{ZIP-FIT} enables the alignment of large-scale datasets with a target domain without the computational burden of neural embeddings.
Our experiments with different compression algorithms (Figure~\ref{fig:compression_comparison}) reveal that lighter compression (e.g., LZ4 at level 0) leads to better performance, achieving a 12.19\% Pass@1 on HumanEval compared to 11.58\% with gzip. This suggests that while compression effectively captures alignment signals, aggressive compression can remove subtle but important patterns. These insights highlight the importance of careful selection of compression parameters in optimizing the quality of data selection.
Our results show that using compression-based alignment leads to faster convergence and lower cross-entropy loss compared to existing methods like DSIR and D4 \citep{d4, dsir}.

However, this approach highlights the trade-off between simplicity and the ability to capture complex semantic relationships. 
While compression-based methods offer a lightweight alternative, they might not fully replace embedding-based techniques for highly intricate domains, such as natural language understanding or paraphrases. 
Nonetheless, \texttt{ZIP-FIT}’s promising results suggest that leveraging compression as a data selection tool can be highly effective, especially in resource-constrained scenarios and economically crucial tasks like code generation, where \texttt{gzip} can leverage the syntactic structure of the data.

Future work could explore hybrid models that combine the strengths of compression-based techniques with neural embeddings to further enhance data selection.
Additionally, extending \texttt{ZIP-FIT} to support more diverse data modalities and investigating its robustness across various domains would provide a more comprehensive understanding of its capabilities and limitations.
We plan for future work to study its application to complex natural language-only tasks and mathematics, where paraphrasing and semantics are important.

We also plan to explore the use of \texttt{ZIP-FIT} for synthetic data generation. While generating synthetic data is straightforward, selecting high-value samples for training presents challenges, especially when managing limited token budgets \cite{villalobos2024rundatalimitsllm}. 
Autoformalization is a fantastic task for this exploration, as it inherently has a limited number of tokens, thus simulating the critical challenge of token scarcity. 
Additionally, studying synthetic data selection is crucial for developing self-improving agents that can avoid model collapse \citep{gerstgrasser2024modelcollapseinevitablebreaking,kazdan2024collapseorthrive} by ensuring high-quality data accumulation.

Furthermore, diversity was identified as an important meta-data property that can influence model performance \citep{miranda2024scalediversitycoefficientdata}. 
Therefore, we aim to address this in future work by either:  
(1) developing an algorithm that balances diversity with alignment in data selection, or 
(2) creating a metric that incorporates diversity as part of its evaluation process.

\begin{mdframed}[backgroundcolor=blue!10, linecolor=blue!50!black, linewidth=2pt, innertopmargin=\baselineskip, innerbottommargin=\baselineskip, innerrightmargin=20pt, innerleftmargin=20pt, roundcorner=10pt]
\textbf{Key Takeaways:}
\begin{itemize}
    \item \textbf{Efficiency in Data Selection:} \texttt{ZIP-FIT} utilizes \texttt{gzip} compression for alignment, demonstrating significant efficiency in selecting domain-specific data, enhancing model fine-tuning.
    \item \textbf{Resource Optimization:} It outperforms traditional methods like DSIR and D4 by speeding up training and reducing computational demands, beneficial in resource-limited settings.
    \item \textbf{Domain-Specific Improvements:} Exhibits superior performance in tasks like AutoFormalization and code generation, where precise data alignment is crucial.
    \item \textbf{Practical Application:} Effective in identifying and using the most relevant data from mixed datasets, proving critical for achieving better domain-specific results.
\end{itemize}
\end{mdframed}

\section{Conclusion}

In this work, we introduced \texttt{ZIP-FIT}, an efficient and scalable data selection method that leverages \texttt{gzip}-based compression to enhance the downstream performance of language models for domain-specific tasks. Our experiments demonstrate that \texttt{ZIP-FIT} not only accelerates the fine-tuning process but also significantly improves downstream performance by aligning training data more closely with target tasks. By comparing against established methods like DSIR and D4, \texttt{ZIP-FIT} proved superior in selecting highly-aligned data, especially in complex tasks such as Autoformalization and code generation. This methodology sets a new standard for resource-efficient and effective data selection for model training, providing a step in understanding the choice of training
data for downstream transfer in LMs.

% \section*{Contributions}

% \section*{Acknowledgments}

\clearpage

\bibliography{iclr2025_conference,references_rylan}
\bibliographystyle{iclr2025_conference}

\clearpage

\appendix

\section{\texttt{gzip} Compression Details}
\label{app:sec:gzip_compression_details}

\texttt{gzip} is a lossless data compression algorithm that combines two primary techniques: LZ77 compression and Huffman coding. Here, we provide additional technical details on how \texttt{gzip} works.

\paragraph{LZ77 Compression:}
LZ77 works by identifying repeated substrings in the input text and replacing them with backward references. Mathematically, LZ77 can be described as follows:

Given an input sequence \( S = s_1, s_2, \dots, s_n \), the algorithm searches for the longest prefix of the remaining sequence \( S' = s_i, s_{i+1}, \dots, s_n \) that matches a substring within a predefined window of previous characters. If a match is found, it is replaced by a tuple \( (d, l, c) \), where:
\begin{itemize}
    \item \( d \) is the distance from the current position to the start of the matching substring,
    \item \( l \) is the length of the matching substring, and
    \item \( c \) is the character following the match (if any).
\end{itemize}
For example, the substring \( s_i, s_{i+1}, \dots, s_{i+l-1} \) can be replaced by the tuple \( (d, l, c) \), thereby reducing redundancy in the data.

\paragraph{Huffman Coding:}
After applying LZ77, \texttt{gzip} employs Huffman coding to further reduce the size of the compressed data. Huffman coding assigns variable-length codes to symbols based on their frequency of occurrence, with shorter codes assigned to more frequent symbols.

The expected length \( L(X) \) of the Huffman code for a sequence of symbols \( X = x_1, x_2, \dots, x_n \) is calculated as:
\[
L(X) = \sum_{i=1}^{n} p(x_i) \cdot \text{len}(C(x_i)),
\]
where:
\begin{itemize}
    \item \( p(x_i) \) is the probability of symbol \( x_i \),
    \item \( \text{len}(C(x_i)) \) is the length of the Huffman code for \( x_i \).
\end{itemize}
This further minimizes the size of the compressed data by leveraging the statistical properties of the input.

\paragraph{Combined \texttt{gzip} Compression:}
The total compressed size \( C(S) \) after applying both LZ77 and Huffman coding can be approximated as the sum of the lengths of the backward references and the Huffman-coded symbols:
\[
C(S) = \sum_{(d, l, c)} \text{len}(d, l, c) + \sum_{i=1}^{n} \text{len}(C(x_i)).
\]

\paragraph{Normalized Compression Distance (NCD):}
\texttt{gzip}’s effectiveness in data selection stems from its ability to measure the alignment between two sequences \( A \) and \( B \) based on how efficiently they compress together. The \textbf{Normalized Compression Distance (NCD)} is given by:
\[
NCD(A, B) = \frac{C(A \oplus B) - \min(C(A), C(B))}{\max(C(A), C(B))},
\]
where \( C(A) \) and \( C(B) \) are the compressed lengths of sequences \( A \) and \( B \), and \( C(A \oplus B) \) is the length of the compressed concatenation of both sequences. 
A lower NCD indicates greater alignment between the sequences.

% If the two objects are very similar, their combined compressed size should approach the size of the simpler one, because you are just adding redundant information.

\subsection{Why Use Compression?}

Compression algorithms, such as \texttt{gzip}, provide a computationally efficient way to detect patterns and minimize redundancy in data. 
\paragraph{Limitations of n-grams:} Many traditional methods, including hashed n-grams, focus on capturing immediate textual correlations by simplifying text into discrete, fixed-size buckets. 
Although these techniques are computationally efficient, they may not adequately capture syntactic or structural relationships within the data. Additionally, the introduce noise due to collisions during hashing.

\paragraph{Challenges with Neural Embeddings:} Neural embeddings offer a powerful tool for capturing semantic relationships, but they come with significant computational costs. These embeddings are typically pre-trained on large corpora and fine-tuned for specific tasks, which requires substantial resources. Given the scalability challenges of embedding-based methods, we conjecture that a simpler method like compression can provide a more scalable and resource-efficient alternative.

We hypothesize that compression -- in this case \texttt{gzip}, but perhaps a different compression algorithm --serves as a strong proxy for capturing syntactic and structural relationships in textual sequences. 
\texttt{gzip}’s ability to compress data based on redundancy minimization can be leveraged as a metric to align text with a target distribution.

\subsection{Composition of the Source Dataset for AutoFormalization}
\label{app:subsec:dataset_composition_af}
The source dataset for the AutoFormalization task was compiled from a variety of datasets to ensure a diverse mix of mathematical, general textual, and code-related content. Below are the details of the datasets included:

\begin{itemize}
    \item \textbf{UDACA/AF:} 4,300 samples from informal formalization statements.
    \item \textbf{C4:} 10,000 samples from the clean crawl of the internet, ensuring a broad linguistic variety.
    \item \textbf{LeanDojo:} 10,000 samples from task-oriented proofs and tactics.
    \item \textbf{LeanDojo Informalized:} 10,000 samples combining traced tactics with informal descriptions, aiming to bridge formal reasoning and natural language.
    \item \textbf{UDACA/AF-split:} 10,000 samples, a variant of the UDACA/AF dataset with split annotations.
    \item \textbf{WikiText:} 10,000 samples from a collection of professionally curated articles, providing a rich linguistic framework.
    \item \textbf{Algebraic Stack:} Samples from various subsets of mathematical and programming languages, capped at 10,000 samples per subset or fewer if the total subset size was under this threshold.
\end{itemize}

Each dataset was selected to complement the others by covering different aspects of language use, from technical to informal, ensuring the model's exposure to a wide range of linguistic structures and contents. The total dataset size aggregated to approximately 185,000 sequences, which were then subjected to alignment scoring and further processing for model training.

\subsection{Composition of the Source Dataset for Code Generation}
\label{app:subsec:dataset_composition_code}

The source dataset for the Code Generation task was assembled from various data sources to provide a diverse range of coding and natural language contexts. Below are the details of the datasets included:

\begin{itemize}
    \item \textbf{MBPP (Google Research):} A total of 964 samples focusing on Python coding challenges.
    \item \textbf{Python Code Instructions (18k Alpaca):} 5,000 sequences providing natural language prompts for Python code, fostering a practical approach to code generation.
    \item \textbf{Python Docstrings (Calum/The Stack):} 5,000 sequences each of Python function docstrings integrating detailed natural language documentation of python functions.
    \item \textbf{Python Docstrings (Calum/The Stack):} 5,000 sequences each of Python function code bodies, integrating raw python code without documentation.
    \item \textbf{C4 (AllenAI):} 10,000 samples from a clean web crawl.
    \item \textbf{WikiText:} 10,000 samples from a collection of curated articles, providing rich natural language training material.
    \item \textbf{Algebraic Stack:} A selection of sequences from various programming language subsets, each capped at 10,000 samples or the total subset size if less than this threshold.
\end{itemize}

This combination of datasets was specifically chosen to challenge our methods 's ability to choose syntactically correct and functionally accurate Python code, while also responding appropriately to natural language prompts. 

\subsection{Hyperparameters for Model Fine-Tuning}

All models in our experiments were fine-tuned with the following unified setup, aimed at ensuring a consistent evaluation across different models and data selection strategies.

\paragraph{Models and Tokenizer:}
The fine-tuning was performed using the following models:
\begin{itemize}
    \item InterLM-Math-Plus-1.8B
    \item Gemma2-2B
    \item Mistral7B
\end{itemize}

\paragraph{Training Settings:}
The key hyperparameters used across all models are as follows:
\begin{itemize}
    \item \textbf{Block Size:} 1024 tokens
    \item \textbf{Learning Rate:} $7.5 \times 10^{-7}$
    \item \textbf{Batch Size:} 4 (per device)
    \item \textbf{Number of Epochs:} 1
    \item \textbf{Weight Decay:} 0.01
    \item \textbf{Maximum Gradient Norm:} 1.0
\end{itemize}

Training was facilitated using the \texttt{Trainer} class from Hugging Face's Transformers library, with the Accelerate library handling model parallelism to efficiently utilize available computational resources.

\paragraph{Evaluation Metrics:}
For model evaluation, we employed:
\begin{itemize}
    \item \textbf{Cross-Entropy Loss} at the end of training to measure the effectiveness of the fine-tuning.
\end{itemize}

Fine-tuning was performed under controlled conditions to ensure fair comparison between data selected by \texttt{ZIP-FIT}, DSIR, and manual curation methods. The effectiveness of each method was assessed based on how the models performed on the ProofNet and HumanEval.

\paragraph{Data Handling and Logging:}
All logs, model checkpoints, and tokenizer settings were systematically saved in designated directories for thorough analysis post-experiment

This comprehensive and standardized approach to fine-tuning ensures that our experimental results are robust, reproducible, and transparent, providing clear insights into the effectiveness of the data selection methodologies employed in our study.

\section{Rationale for the Method Name \texttt{ZIP-FIT}}

We chose the name \textbf{ZIP-FIT} for two reasons:
\begin{enumerate}
    \item \textbf{ZIP} refers to the use of \texttt{gzip} compression for data selection, where compression aligns the data for better future fine-tuning (or \textbf{FITting}).
    \item The name also references scaling laws, as \texttt{ZIP-FIT} consistently reduces loss faster than competing methods, implying better power-law scaling parameters, drawing a parallel to \textbf{Zipf's law} \cite{piantadosi2014zipf}, which describes similar scaling behavior in language models.
\end{enumerate}

\textit{Remark:} Zipf's law \cite{piantadosi2014zipf} describes the inverse relationship (thus power law $f(r) \propto 1/r^s$, where $r$ is the rank and $f(r)$ is the frequency of the word with rank $r$) between a word's frequency and its rank in natural language, a pattern that reflects scaling behavior. 
Rank in this context is the position of the word after sorting with respect to frequency in the text.

\newpage

\section{Additional Experimental Results: data selection for efficient fine-tuning using \texttt{ZIP-FIT}}
\label{sec:more-af-plots}
\begin{figure*}[!ht]
    \centering
    \includegraphics[width=\textwidth]{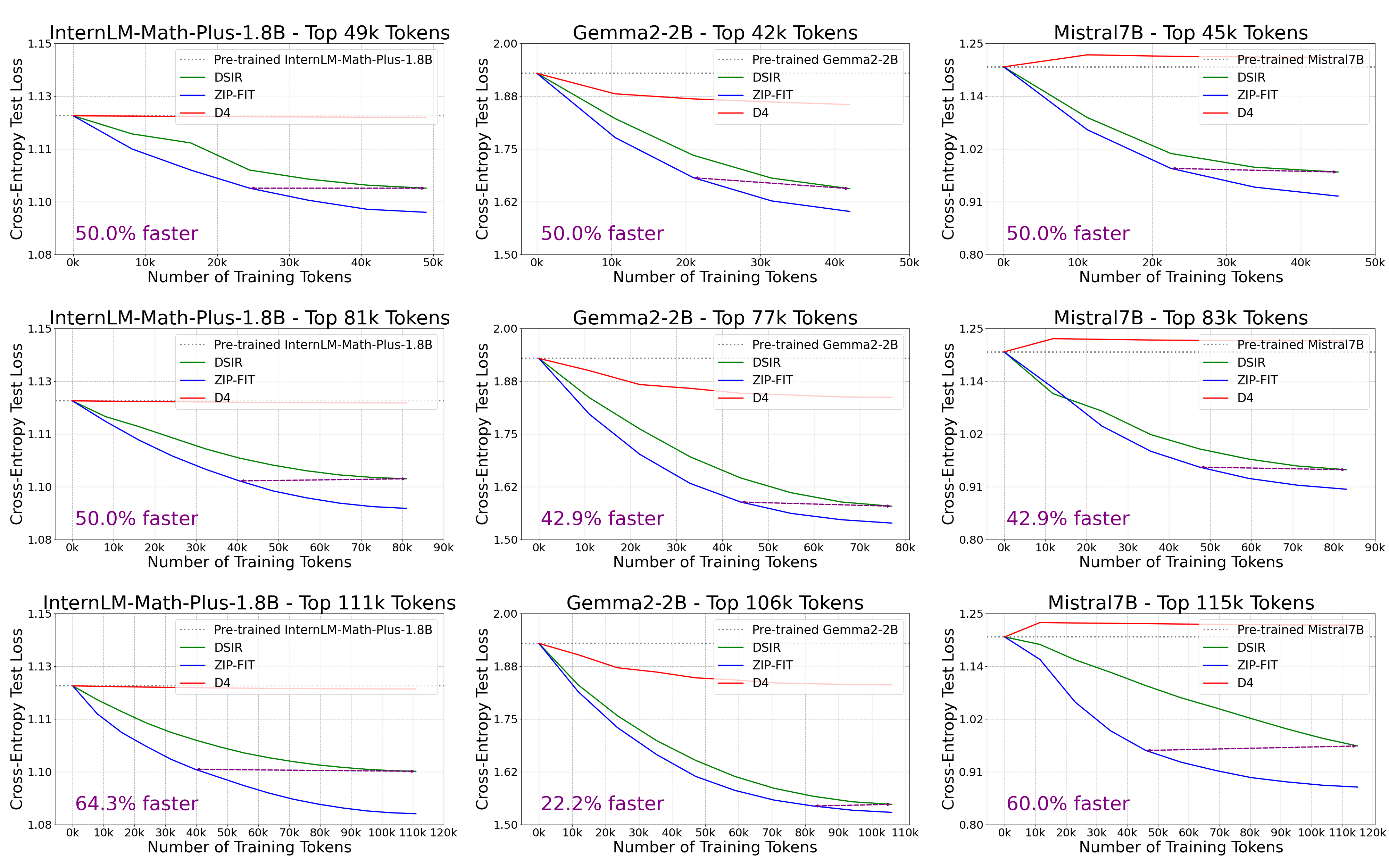}
    \caption{\textbf{\texttt{ZIP-FIT} consistently achieves a lower test loss at a faster rate compared to D4 and DSIR for Autoformalization}. The plots show the cross-entropy test loss against the number of training tokens for three models (InterLM-Math-Plus-1.8B, Gemma2-2B, and Mistral7B) across various token selection sizes. \texttt{ZIP-FIT} (blue line) consistently surpasses both DSIR (green line) and D4 (red line) across all model and token size configurations, emphasizing its superior data processing efficiency. The percentage labels in each plot denote the relative speedup of \texttt{ZIP-FIT} over DSIR in attaining the lowest cross-entropy loss, further underscoring the method’s scalability and adaptability for domain-specific fine-tuning.}
    \label{fig:af_all_models_lower_ks}
\end{figure*}

\newpage

\section{Baseline Comparison using Pass@1 on HumanEval}
ZIP-FIT demonstrates substantial improvements in code generation capability across different fine-tuning approaches. When applied to 4-bit quantized models with LoRA fine-tuning, ZIP-FIT doubles the Pass@1 performance of the base model and outperforms existing data selection methods. 

\begin{table}[!h]
\centering
\caption{\textbf{Performance and efficiency comparison of data selection methods.} Results show Pass@1 and Pass@10 scores on HumanEval using top 1M tokens for fine-tuning, along with data selection time. Data selection times exclude fine-tuning time.}
\begin{tabular}{llccc}
\toprule
\textbf{Fine-tuning} & \textbf{Data Selection} & \textbf{Pass@1 (\%)} & \textbf{Pass@10 (\%)} & \textbf{Selection Time} \\
\midrule
None & Pre-trained Gemma2-2B & 15.24 & 38.81 & -- \\
None & Pre-trained Gemma2-2B (4-bit quantized) & 6.09 & -- & -- \\
\midrule
Full FT & ZIP-FIT & \textbf{18.86} & 41.78 & \textbf{32s} \\
Full FT & LESS & 18.06 & 40.19 & 19h \\
Full FT & DSIR & 17.98 & \textbf{44.27} & 97s \\

\midrule

QLoRA & ZIP-FIT & \textbf{12.19} & -- & \textbf{32s} \\
QLoRA & DSIR & 9.14 & -- & 97s \\
QLoRA & D4 & 6.09 & -- & 7h 40m \\
\bottomrule
\end{tabular}
\end{table}

\newpage

\section{Impact of Compression Algorithms and Levels}

\begin{figure}[h]
\centering
    \includegraphics[width=\textwidth]{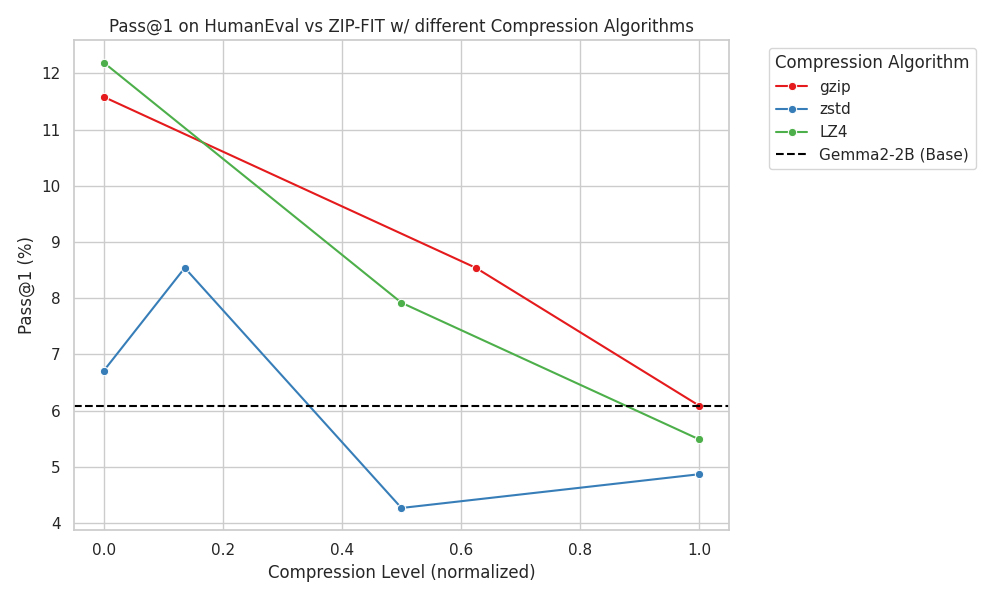}
\caption{\textbf{Lighter compression preserves crucial information for data selection.} At minimum compression levels, both gzip and LZ4 achieve the strongest Pass@1 scores (11.58\% and 12.19\%), significantly outperforming the base model (6.09\%, dashed line). Performance systematically degrades with increased compression across all algorithms, suggesting that aggressive compression removes valuable alignment signals.}

\label{fig:compression_comparison}
\end{figure}

To investigate the impact of different compression algorithms on ZIP-FIT's performance, we conducted experiments comparing three widely used compression methods: gzip, zstd, and LZ4. Each algorithm was tested across its available compression levels, normalized to a 0-1 scale for comparison. As shown in Figure~\ref{fig:compression_comparison}, compression algorithm choice and level significantly impact performance.

Key findings include:
\begin{itemize}
    \item LZ4 at minimum compression achieves the best performance (12.19\% Pass@1)
    \item Higher compression levels generally lead to decreased performance across all algorithms
    \item gzip shows more stable performance degradation compared to LZ4 and zstd
    \item zstd consistently underperforms relative to both GZIP and LZ4
\end{itemize}

These results suggest that lighter compression better preserves the structural information needed for effective data selection. The superior performance of LZ4 at minimal compression indicates that aggressive data compression may remove subtle but important patterns useful for alignment assessment.

\newpage

\section{Data Selection Profiling (Run Times)}\label{sec:profiling_run_times}
\begin{figure*}[!ht]
    \centering
    \includegraphics[width=0.8\textwidth]{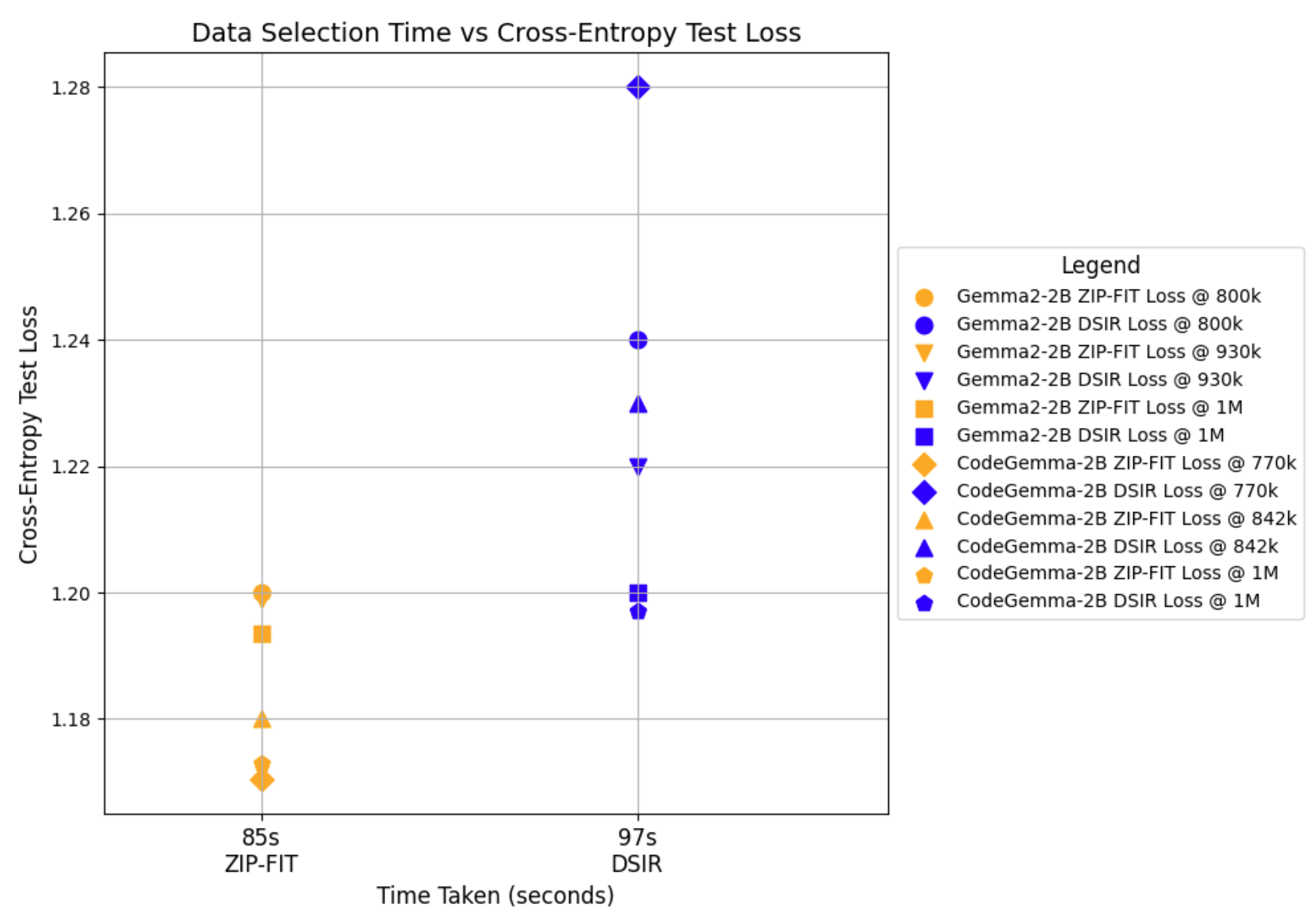}
    \caption{\textbf{\texttt{ZIP-FIP} demonstrates lower cross-entropy and lower run time during data selection than competing DSIR and D4 methods.} 
    \texttt{ZIP-FIT} is cheaper, faster, and better performing. 
    The run times do no include fine-tuning time, since it’s a constant offset across all models. 
    D4's data selection (not shown) takes 5hs because it uses an embedding model (opt-125m \cite{opt}), the same one as the original paper \cite{d4}.
    }
    \label{fig:profiling_run_times}
\end{figure*}

\texttt{ZIP-FIT} performs selection up to 65.8\% faster than DSIR and 21,076\% (=5h/85s=211, which is 2 orders of magnitude) faster than D4. 
Experimental results comparing \texttt{ZIP-FIT} vs DSIR profiling/run time for Code data selection can be found in figure \ref{fig:profiling_run_times}. Note that depending on the dataset and number of samples these numbers may not hold. Compression may not scale well to long-context datasets and depending on the source dataset, our run times varied widely. However, on average we observed that \texttt{ZIP-FIT} is comparable to DSIR and generally faster. More experiments across a wider range of datasets need to be conducted in order to infer more.

\newpage

\section{Qualitative Analysis}
Qualitative results show top 20 examples can be found it table \ref{tab:code_top_20_zip_fit}.

\begin{table}[htbp]
    \centering
    \textbf{Selected Samples by \texttt{ZIP-FIT} with \texttt{ZIP-FIT} Alignment scores}\\[0.5em] 
    \begin{tabular}{|p{10cm}|c|}
        \hline
        \textbf{Sample Text (Beginning)} & \textbf{Alignment Score} \\ \hline
        Across all his bands and projects, Townsend has released twenty @-@ three studio albums and three live albums. & 0.5000 \\ \hline
        Require Import CodeDeps. Require Import Ident. Local Open Scope Z\_scope. Definition \_addr := 1\%positive. Definition \_g := 2\%positive. & 0.4928 \\ \hline
        This Photostock Vector Night Sky Background With Full Moon Clouds And Stars Vector Ilgraphicration has 1560 x 1560 pixel resolution... & 0.4926 \\ \hline
        module Structure.Logic where ... & 0.4926 \\ \hline
        \{ dg-do compile \} PR fortran/51993 Code contributed by Sebastien Bardeau \texttt{<bardeau at iram dot fr>} module mymod type :: mytyp... & 0.4891 \\ \hline
        For over ten years, the St. Louis Mercy home has formed a special connection with a local community theatre: The Muny. This summer the... & 0.4889 \\ \hline
        Read("SchreierSims.gi"); LoadPackage("AtlasRep"); MicroSeconds := function() local t; t := IO\_gettimeofday(); return t.tv\_sec * 1000000 + t.t & 0.4889 \\ \hline
        Get the keyId used by this peer (this peer's identifier). This is stored in the key store. & 0.4857 \\ \hline
        Initializes and adds a node to the graph. NOTE: At least the type must be supplied for the Node to exist in the graph. Args: graph: The graph... & 0.4853 \\ \hline
        def bgra2rgb(img): cv2.cvtColor(img, cv2.COLOR\_BGRA2BGR) has an issue removing the alpha channel, this gets rid of wrong trans... & 0.4853 \\ \hline
        % \texttt{\{\# LANGUAGE Strict, StrictData \#\} \{\# LANGUAGE DeriveGeneric, DeriveAnyClass \#\} \{\# LANGUAGE TypeFamilies, TypeFamilyDependencies \#\}...} & 0.4853 \\ \hline
        % \#include \texttt{<gsl/gsl\_math.h>} \#include "gsl\_cblas.h" \#include "cblas.h" void cblas\_drotmg (double *d1, double *d2, double *b1... & 0.4818 \\ \hline
        % Minnesota Starvation Experiment: To study the effects of diet and nutrition, Dr. Ancel Keys of the University of Minnesota Laboratory of Physiologi... & 0.4815 \\ \hline
        % \#include \texttt{<empool/ThreadPoolManager.h>} \#include \texttt{<empool/Mutex.h>} \#include \texttt{<empool/TaskBase.h>} \#include \texttt{<atomic>}... & 0.4815 \\ \hline
        % It’s in the details of 100,000 moments. What made you happy in the past 24 hours? Researchers asked 10,000 people this question... & 0.4779 \\ \hline
        % def run(*args: Any, **kwargs: Any) -> None: 'Run cwltool.' windows\_check() signal.signal(signal.SIGTERM, \_signal\_handler) try: & 0.4714 \\ \hline
        % Args: backbone: either a backbone module or a mmdet config dict that defines a backbone. The backbone takes a 4D image tensor and returns... & 0.4706 \\ \hline
        % \{ SPDX-License-Identifier: MIT \} \#include \texttt{<armnn/Descriptors.hpp>} \#include \texttt{<armnn/IRuntime.hpp>}... & 0.4706 \\ \hline
        % \{\# OPTIONS --no-positivity-check \#\} module STLC.Coquand.Completeness where open import STLC.Coquand.Normalisation public... & 0.4672 \\ \hline
        % // Copyright \texttt{©} 2017 Arm Ltd. All rights reserved. SPDX-License-Identifier: MIT & 0.4667 \\ \hline
    \end{tabular}\label{tab:code_top_20_zip_fit}
    \caption{Beginning characters of the top 20 samples selected by \texttt{ZIP-FIT} when the target task is code generation.}
\end{table}

\begin{table}[htbp]
    \centering
    \textbf{Selected Samples by DSIR with \texttt{ZIP-FIT} Alignment scores}\\[0.5em] 
    \begin{tabular}{|p{10cm}|c|}
        \hline
        \textbf{Sample Text (Beginning)} & \textbf{\texttt{ZIP-FIT} Alignment Score} \\ \hline
        \textless a href=``https://colab.research.google.com/github/julianovale/simulaca
        o\_python/blob/master/0006\_ex\_trem\_kronecker\_algebra\_computacao ... & 0.122 \\ 
        \hline
        library(qcc) \textbackslash\textbackslash n
        death=c(2,1,2,4,2,5,3,3,5,6,3,8,3,3,6,3,6,5,3,5,2,6,2,3,4,
        3,2,9,2,2,3,2,10,7,9,6,2,1,2,4,2,5,3,3,5,6,3,8,3,3,6,3,6,5,3,5,2,6,2 ... & 0.121 \\ 
        \hline
        gap \textgreater List(SymmetricGroup(4), p - \textgreater Permuted([1 .. 4], p)); \textbackslash\textbackslash n perms(4);
        [ [ 1, 2, 3, 4 ], [ 4, 2, 3, 1 ], [ 2, 4, 3, 1 ], [ 3, 2, 4, 1 ... & 0.191 \\
        \hline
        \# Solutions \textbackslash\textbackslash n
        \#\# Question 1 \textbackslash\textbackslash n
        \textgreater `1'. Using a `for' loop print the types of the variables in each of the
        \textgreater following iterables: \textbackslash\textbackslash n
        \textgreater  `1' ... & 0.145 \\
        \hline
        \# Some small pregroups \textbackslash\textbackslash n \# The lists of small pregroups were generated by \textbackslash\textbackslash n
        \# Chris Jefferson <caj21@st-andrews.ac.uk> and \textbackslash\textbackslash n ... & 0.195 \\ 
        \hline
        adjacency\_mat = [
        false true true true true true true true true false true true true true false 
        false false true false true false true false ... & 0.182 \\
        \hline
        \textbackslash section\{Lookup table used for accessing child voxels using a parent's 
        child descriptor\}
        \textbackslash label\{app:lookup-table\}
        \textbackslash lstset\{language=C,cap ... & 0.199 \\
        \hline
        \textbackslash* statistics \ test\_nist.c \textbackslash\textbackslash n * \textbackslash\textbackslash n * Copyright (C) 1996, 1997, 1998, 1999, 2000, 2007 Jim Davies, Brian Gough         \textbackslash\textbackslash n *         \textbackslash\textbackslash n * This pro ... & 0.180 \\
        \hline
        Problem Description Write a python function to find the first missing positive number. \textbackslash\textbackslash n def first\_Missing\_Positive(arr,n): \textbackslash\textbackslash n ptr = 0 ... & 0.239 \\
        \hline 
        import numpy as np \textbackslash\textbackslash n mandelTable = [[0,0,0,0,0,0,0,0,0,0,0,0,0,0,0,0,0,0,0,0,0,0,0,0,0,0,0,0,0,0,0,0,0, ... & 0.189 \\
        \hline
    \end{tabular}\label{tab:code_top_20_dsir}
    \caption{Beginning characters of the top 20 samples selected by DSIR when the target task is code generation. 
    DSIR does not easily provide alignment scores, so instead we report the \texttt{ZIP-FIT} scores, which reveals that \texttt{ZIP-FIT} doesn't score highly the DSIR examples which might explain why \texttt{ZIP-FIT} achieves better CE loss. 
    }
\end{table}

\section{Future Work (Cont.)}

\textbf{Lossless Compression for Alignment:} While \texttt{ZIP-FIT} has demonstrated substantial efficiency for data selection, there are several promising directions for future exploration. 
One potential enhancement is leveraging faster compression algorithms, such as \texttt{LZ4} and \texttt{Snappy}, which offer rapid processing speeds at the cost of lossy compression. 
In our current approach, we utilize \texttt{gzip} for compression-based alignment, which is lossless and provides a robust foundation. 
However, \texttt{LZ4} and \texttt{Snappy} are optimized for speed and could potentially offer even greater computational efficiency without the need for decompression in our pipeline. 
Given that our primary goal is efficient data selection rather than perfect data recovery, these faster algorithms might be more suitable. 

\textbf{Autonomous Validation Set Generation}: 
A current limitation of ZIP-FIT is its dependence on a small, curated validation set (e.g., 185 samples for ProofNet and 82 samples for half the HumanEval test set). 
Future work could explore the use of generative models to create synthetic validation sets from task-specific instructions. 
This approach could also be expanded to enable autonomous self-directed, model-driven generation of validation data.

\end{document}